\documentclass{llncs}

\usepackage[a4paper,includeheadfoot,top=1.65in,bottom=1.65in,left=1.73in,hcentering]{geometry}
\usepackage[pdftex]{graphicx}
\usepackage[sectionbib]{natbib}

\usepackage[utf8]{inputenc} 
\usepackage{t1enc}
\usepackage{tikz-dependency}
\usepackage{qtree}
\usepackage{float}

\usepackage{graphicx}%
\usepackage{multirow}%
\usepackage{amsmath,amssymb,amsfonts}%
\usepackage{mathrsfs}%
\usepackage[title]{appendix}%
\usepackage{xcolor}%
\usepackage{textcomp}%
\usepackage{manyfoot}%
\usepackage{booktabs}%
\usepackage{algorithm}%
\usepackage{algorithmicx}%
\usepackage{algpseudocode}%
\usepackage{listings}%
\usepackage{comment}
\usepackage{hhline}
\usepackage{todonotes}
\usepackage{comment}
\usepackage{hyperref}  
\usepackage{siunitx}
\ifx\anonym\undefined
  \newcommand{\racka}{Racka}
\else
  \newcommand{\racka}{OurModel}
\fi

\restylefloat{figure}
\usepackage{tikz-qtree}
\usepackage[english]{babel}
\newcommand{\furl}[1]{\footnote{\url{#1}}}
\newcommand{\fhref}[2]{\href{#1}{#2}\footnote{\url{#1}}}

\begin{document}

\title{\racka{}: Efficient Hungarian LLM Adaptation on Academic Infrastructure}
\author{
\ifx\anonym\undefined
    Zsolt Csibi\inst{2}\thanks{Authors in alphabetical order.},
    Bence Gy\"orgy Gortka\inst{1},
    Natabara Gy\"ongy\"ossy\inst{2},
    Korn\'el Nagy\inst{1},
    D\'avid M\'ark Nemeskey\inst{1},
    G\'abor Palk\'o\inst{1},
    Martin Sallai\inst{1},
    Andr\'as Simonyi\inst{2},
    Andr\'as M\'ark Szekeres\inst{1}
\fi
}

\institute{
\ifx\anonym\undefined
    $^1$ELTE Faculty of Humanities, Department of Digital Humanities\break
    \{gortka.bence, nagy.kornel, nemeskey.david, palko.gabor, sallai.martin, szekeres.andras.mark\}@btk.elte.hu\break
    $^2$ELTE Faculty of Informatics, Department of Artificial Intelligence\break
    \{vrnf2j, natabara, simonyi\}@inf.elte.hu
\fi
}

\maketitle

\begin{abstract}
We present \racka{}-4B, a lightweight, continually pretrained large language model designed to bridge the resource gap between Hungarian and high‑resource languages such as English and German. \racka{} employs parameter‑efficient continual pretraining via Low‑Rank Adaptation (LoRA) on a Qwen3-4B backbone, making the recipe practical on A100 (40GB)‑based HPC clusters with low inter‑node bandwidth. To better match the training distribution, we replace and adapt the tokenizer. By swapping 32k tokens, we achieve 47\% lower tokenization fertility and generation latency for Hungarian while maintaining competitive performance in English and German. The model is trained on 160B subword tokens (85B whitespace words) drawn from a mixture of internet and high‑quality curated sources, with a composition of 44\% Hungarian, 24\% English, 21\% German, and 11\% code to mitigate catastrophic forgetting. Our evaluations indicate modest but stable results in language adaptation with similar LM-Eval-Harness results but increased overall performance on HULU and OpenHuEval compared to the original Qwen and the latest Puli models. The results also showcase that \racka{}-4B is capable of Hungarian chat with English reasoning even in the absence of explicit Hungarian post-training on these tasks. 
\end{abstract}

\section{Introduction}
\label{sec:intro}

\subsection{The Challenge of Linguistic Digital Sovereignty for Medium-Resource Languages}
The proliferation of Large Language Models (LLMs) has marked a transformative era in artificial intelligence, yet this progress has been unevenly distributed across the world's languages~\citep{ali2024surveylargelanguagemodels}. The current landscape is dominated by models pretrained on vast corpora composed predominantly of English and a few other high-resource languages, creating a significant performance and resource disparity for less-resourced linguistic communities~\citep{zhong2025opportunitieschallengeslargelanguage}. For medium-resource languages such as Hungarian, a Finno-Ugric language characterized by its agglutinative nature and rich morphology, this gap is particularly pronounced. Off-the-shelf multilingual models often exhibit suboptimal performance due to insufficient representation in training data and tokenizers that are ill-suited to language-specific morphology. This is particularly the case for open-source models, which visibly struggle with Hungarian grammar.

This disparity highlights a critical challenge to linguistic digital sovereignty. Linguistic communities need the capacity to develop, deploy, and benefit from AI technologies that are not only proficient in their language but are also attuned to their unique cultural and contextual nuances. Achieving this requires the creation of models trained on nationally relevant data and, crucially, developed using methodologies that are accessible to local research communities. Likely, most research groups will not have access to the hyperscale computing infrastructure available to large industrial labs and will have to rely on shared, and often underpowered, HPC infrastructure.

\subsection{Core Contributions}
In response to these challenges, we introduce \racka{}, a 4-billion-parameter language model for Hungarian adapted trough a pragmatic and resource-efficient approach from a Qwen3-4B reasoning model.
\ifx\anonym\undefined{
  The model's name pays homage to the \textit{Hortobágy Racka}, an indigenous Hungarian sheep breed and llama alternative.
\fi
\racka{} was trained using publicly accessible infrastructure, Hungary's Komondor supercomputer, rather than state-of-the-art proprietary hardware.
Our primary contributions are as follows:


\begin{enumerate}
    \item \textbf{Curation of a Large-Scale Multilingual Corpus:} We describe the assembly and curation of a 160 billion subword token dataset representing state-of-the-art size in training a Hungarian-centric language model. We justify its strategic composition (44\% Hungarian, 43\% high-resource languages, and 11\% code) as a pragmatic approach to mitigate catastrophic forgetting during the continual pretraining process. The inclusion of English and German is justified by the census report of the Hungarian Central Statistical Office\footnote{\url{https://nepszamlalas2022.ksh.hu/eredmenyek/vegleges-adatok/kiadvany/}}, as these are the most common foreign languages in Hungary. 
    \item \textbf{Optimized Tokenization for Hungarian:} We underscore the critical role of tokenizer adaptation for morphologically rich. We document the design, training, and evaluation of a new tokenizer that dramatically improves tokenization efficiency (i.e., subword fertility) for Hungarian, while carefully preserving the performance of the base model on high-resource control languages like English and German.
    \item \textbf{A Resource-Efficient Training Recipe:} We detail a practical methodology for the continual pretraining of a 4B parameter model on an HPC cluster featuring NVIDIA A100 (40GB) GPUs. We provide justification for our selection of Distributed Data Parallel (DDP) over the more recent Fully Sharded Data Parallel (FSDP) paradigm, demonstrating that this choice is a deliberate optimization for our specific hardware environment.
    \item \textbf{The first Hungarian reasoning LLM:} We present \racka{}, which to our knowledge is the first Hungarian LLM with reasoning output. The model is openly available on Hugging Face as \fhref{https://huggingface.co/elte-nlp/Racka-4B}{\texttt{elte-nlp/Racka-4B}}.
\end{enumerate}

\section{Related Work}
\label{sec:related}

\subsection{Paradigms of Language Model Adaptation}
Adapting existing foundation models to new languages, domains, or tasks is a central challenge in modern NLP. The prohibitively high cost of training LLMs from scratch has spurred research into more efficient adaptation strategies~\citep{csaki2023efficientlyadaptingpretrainedlanguage,khade2024challengesadaptingmultilingualllms}. As an industry standard practice for language adaptation, continual pretraining (CPT) is widely adopted. This method focuses on further self-supervised training on large, unlabeled corpora to expand a model's foundational knowledge~\citep{vanroy2024geitje7bultraconversational,laion2024leolm,ke2023continual}.

The work on \racka{} falls squarely within the CPT paradigm, specifically targeting what the literature defines as ``Language Expansion'' and ``Domain Adaptation''. This framing positions our effort not as the creation of a final, task-specific model, but as general fine-tuning of a foundational model for Hungarian, which can subsequently be adapted for downstream tasks. A primary obstacle in any continual learning scenario is \textit{catastrophic forgetting}, where a model's performance on previously learned tasks or languages degrades significantly as it learns new 
information~\citep{ke2023continual}. A simple yet effective strategy to mitigate this is data replay, which involves mixing data from the original training distribution with the new data. Our multilingual data composition, which retains a significant portion of English and German text, is a direct application of this principle. Other alternatives include careful train scheduling, intense regularizations, or the analysis of task-vector shifts. We refrain from these methods in favor of data replay, as it directly controls our training objective and does not require extensive experimentation~\citep{li2024examiningforgettingcontinualpretraining}. Previous Hungarian language adaptation work also applied this approach successfully~\citep{szentmihalyi2024pretraining, csaki2024sambalingoteachinglargelanguage, gyozo2025pulillumix}.

\subsection{Parameter-Efficient Methods for Continual Pretraining}
The computational burden of updating the billions of parameters in an LLM has led to the development of Parameter-Efficient Fine-Tuning (PEFT) methods. These techniques aim to adapt models by training only a small fraction of their total parameters, drastically reducing memory and compute requirements. Among the most widely adopted PEFT methods is Low-Rank Adaptation~\citep[LoRA;][]{hu2021loralowrankadaptationlarge}. LoRA operates on the hypothesis that the change in model weights during adaptation has a low intrinsic rank. This aligns well with our aim to train on limited-memory GPUs with low throughput internode connections.
By keeping the base model frozen, it is shown to inherently aid in overcoming catastrophic forgetting of the base model's knowledge thus making it ideal for CPT~\citep{li2024examiningforgettingcontinualpretraining}.

\subsection{The Landscape of Hungarian Language Models}
The development of \racka{} builds upon a growing body of work dedicated to creating capable language models for Hungarian. These efforts span both academic and industrial research and provide essential context for our contributions.
\begin{itemize}
    \item \textbf{NYTK Models:} The Hungarian Research Centre for Linguistics (NYTK) has been a key contributor, developing a series of models within the ``Puli'' family. These include models in the range of 7-8B parameters, both monolingual and continually pretrained. Latest models include instruction-tuned and multilingual (English, Chinese) models~\citep{yang-puli-gptrio,gyozo2025pulillumix,yang-llumix-llama}.
    \item \textbf{OTP Bank Models:} A major government-supported initiative, lead by OTP Bank, demonstrated the feasibility of developing high-performance models for Hungarian at a large scale~\citep{szentmihalyi2024pretraining}. The resulting models were trained on an extensive dataset on custom SambaNova hardware. Unfortunately, these models are not available to the public.
    \item \textbf{International Initiatives:} The ecosystem also includes models adapted from multilingual open-source foundations. SambaLingo-Hungarian-Base, for instance, adapted a Llama-2 7B model by training on 59 billion tokens from the Hungarian portion of the Cultura-X dataset~\citep{csaki2024sambalingoteachinglargelanguage}. Similarly, projects like OpenEuroLLM have fine-tuned models like Gemma for improved Hungarian performance. Other European initiatives such as EuroLLM and Salamandra include Hungarian in their supported languages. \citep{openeurollm2025hungarian,eurollm2025,gonzalezagirre2025salamandratechnicalreport}
\end{itemize}

This existing landscape highlights a clear and valuable niche for the \racka{} project. While previous efforts have focused on larger models (OTP-13B), monolingual training (PULI-GPT-3SX), multilingual training (PULI-GPTrio) or adaptation on smaller datasets (SambaLingo), \racka{}'s contribution is its unique combination of a lightweight (4B) backbone, continual pretraining on an exceptionally large multilingual corpus (160B tokens), and a resource-efficient methodology explicitly designed for and validated on publicly accessible national HPC infrastructure.
One key insight that can be accrued from the papers above is that for smaller languages, continual pretraining of already capable models results in better performing LLMs than training new models from scratch.

\section{Training Data Curation}
\label{sec:data}

\subsection{Corpus Design and Composition}
The foundation of any successful language model is the quality and scale of its training data. For \racka{}, we curated a comprehensive multilingual corpus totaling approximately 160 billion subword tokens with a dataset size of $639$ GB. The composition of this corpus was a strategic design choice aimed at achieving robust language adaptation for Hungarian while simultaneously mitigating the effects of catastrophic forgetting on the backbone model's existing capabilities. The final data mixture, detailed in Table \ref{tab:corpus-composition}, allocates the majority of the data to our target language, Hungarian, while retaining substantial portions of high-quality English, German, and source code data.
The code component was included to maintain the model's strong reasoning and structured-text processing abilities, which are known to be beneficial for a wide range of downstream analytical tasks~\citep{ma2024at}.

\begin{table}[h!]
\centering
\caption{Composition of the \racka{} Pretraining Corpus after subset weighting and tokenization of the Hungarian-adapted tokenizer.}
\label{tab:corpus-composition}
\setlength{\tabcolsep}{6pt} 
\begin{tabular}{lccc}
\toprule
\textbf{Language} & \textbf{BPE Tokens (Billions)} & \textbf{Ratio} & \textbf{Documents (Millions)} \\ \midrule
Hungarian & $\sim$70 & 44\% & 70.56 \\
English   & $\sim$38 & 24\% & 26.12 \\
German    & $\sim$34 & 21\% & 24.06 \\
Code      & $\sim$18 & 11\% & 10.17 \\ \midrule
\textbf{Total}    & \textbf{$\sim$160} & \textbf{100\%} & \textbf{130.91} \\ \bottomrule
\end{tabular}
\end{table}

\subsection{Data Sources}
The corpus was assembled from a combination of large-scale web crawls and high-quality curated datasets to ensure both breadth and quality. The primary source for web data across all languages was the Common Crawl repository, with all data sourced from crawls conducted prior to our cutoff date of October 2024~\citep{commoncrawl2025}.

For English, we sample approximately $38$B tokens sourced from non-web text-based subsets of The Pile~\citep{gao2020pile800gbdatasetdiverse}, and the heavily filtered FineWeb corpus~\citep{penedo2024finewebdatasetsdecantingweb}.
We sample approximately $34$B tokens of German text from the similarly pre-filtered occiglot-fineweb~\citep{occiglotfineweb2024} and $18$B tokens of code from The Stackv2~\citep{lozhkov2024starcoder2stackv2}, slightly oversampling structured text formats and markups to better represent them compared to code files.
$70$B Hungarian tokens were collected from various sources including Common Crawl, academic repositories, court rulings, movie subtitles, Wikipedia and digital news articles. Following \citet{Nemeskey:2020}, the web and news subcorpora were heavily filtered and deduplicated on both the URL and document-level. Paragraphs that appeared frequently across the respective subcorpora were also discarded. The aggregated per-language corpus sizes are reported in Table~\ref{tab:corpus-composition}; the composition of the Hungarian corpus is detailed in Appendix~\ref{app:hun_corpus}. Note that the fertility of our adapted tokenizer is smaller for Hungarian, thus we produce less tokens in Hungarian compared to the other subsets.

In case of Hungarian web data, we apply standard boilerplate filtering and a strict paragraph-level 13-gram (whitespace token) deduplication resulting in our final dataset size. To ensure that the prevalence of scarce but high quality data is higher in the training dataset, we oversample the news subset (weight $2.0$), the Wikipedia subset (weight $3.0$), Hungarian subtitles (weight $2.0$) and academic repositories (weight $1.5$). We drop those documents that are less than $500$ characters long but do not apply other 
length-based filtering.

For measuring training efficiency we hold out 10-10 thousand element validation and test sets of independent documents. Both sets are sampled in a language-stratified manner.

\section{Methodology}
\label{sec:method}

\subsection{Backbone Model Architecture}
The foundation for our work is the Qwen3-4B model, an open-source transformer-based LLM developed by Alibaba Cloud accessed through Hugging Face\footnote{\url{https://huggingface.co/Qwen/Qwen3-4B}}. We selected this model as our backbone due to its strong baseline performance, favorable open-source license (Apache 2.0), and an architecture well-suited for our objectives. The Qwen3-4B model features $4$ billion total parameters ($3.6$ billion non-embedding) distributed across $36$ transformer layers~\citep{yang2025qwen3technicalreport}.

Key architectural specifications include the use of Grouped Query Attention (GQA) 
for efficient inference, the SwiGLU activation function for improved performance, and RMSNorm for stable training~\citep{yang2025qwen3technicalreport}. The model supports a native context length of 32,768 tokens, making it capable of processing long documents without truncation, which could be extended via interpolation of the positional encoding~\citep{su2023roformerenhancedtransformerrotary}. We specifically chose the reasoning and instruction-tuned variant of the model, hoping that its inherent capabilities would provide a stronger starting point and ultimately benefit in downstream performance after our continual pretraining phase. To our knowledge, this makes \racka{} the first reasoning model to be adapted to Hungarian.

\subsection{Tokenizer Adaptation for Hungarian}
A key hypothesis of our work is that the default tokenizer of a multilingual model is suboptimal for a morphologically rich, agglutinative language like Hungarian. Such tokenizers, trained on a distribution dominated by English, tend to fragment Hungarian words into numerous, less meaningful subword units, leading to longer sequence lengths, increased computational cost, and potentially poorer model performance~\citep{csaki2023efficientlyadaptingpretrainedlanguage}. Several previous models have also experimented with tokenizer adaptation. To address this issue, we developed a new, adapted tokenizer for \racka{}, slightly modifying previously used adaptation algorithms. The process involved several steps:
\begin{enumerate}
    \item We first trained a new Byte-Pair Encoding (BPE) tokenizer from scratch, using only a small portion of our Hungarian training corpus. This produced a vocabulary optimized for the statistical properties of Hungarian.
    \item We then merged this new Hungarian-specific vocabulary with the original vocabulary of the Qwen3 tokenizer. Rare, typically non-latin character-based multi-byte tokens were removed, and novel tokens were inserted resulting in an adapted vocabulary that retained all of the original latin alphabet tokens while adding $32\ 000$ new, Hungarian-optimized tokens learned from the Hungarian-only tokenizer. See Appendix~\ref{app:huntok} for details.
    \item The model's architecture was updated to accommodate this change: the token embedding matrix and the language-model head (the final output layer) were altered to match the new tokenizer. To provide a strong initialization for the newly added tokens in the embedding matrix, we employed the VIPI (Vocabulary Initialization with Partial Inheritance) vocabulary-transfer technique~\citep{mosin2022finetuning}. For each new token, we used the original tokenizer to break it down into its constituent subwords. The new token's initial embedding vector was then calculated as the average of the embedding vectors of these constituent subwords from the original, pretrained embedding matrix. This approach provides a semantically meaningful starting point for the new tokens, facilitating faster convergence during training.
\end{enumerate}

\subsection{Packed Continual Pretraining with LoRA}

To enable parameter-efficient continual pretraining, we employed Low-Rank Adaptation (LoRA). LoRA modules were inserted into all linear layers of the Qwen3-4B backbone. The weights of the base model were frozen throughout the training. The embedding and language modeling head layers are tied as in Qwen3-4B.

Our choice of LoRA hyperparameters was guided by recent best practices and empirical validation. Specifically, we set the LoRA rank ($r$) to $64$ and the scaling factor ($\alpha$) to $128$, following the heuristic of setting $\alpha$ to approximately twice the rank to ensure sufficient adaptation capacity.  A dropout of $0.1$ was applied to the LoRA adapters to improve generalization. With this setup we train $0.52$B parameters accounting for $12.5\%$ of the base model size. We used separate learning rates for LoRA and non-LoRA parameters: $1\times 10^{-4}$ for LoRA and $5\times 10^{-5}$ for non-LoRA parameters, both with a weight decay of $0.005$. Optimization was performed using AdamW. The learning rate schedule included a linear warmup phase covering $1\%$ of total training steps, followed by linear decay to zero which is shown to be beneficial~\citep{bergsma2025straightzerolinearlydecaying}. 
We did not employ a specialized tokenizer healing phase.

To maximize training efficiency, we used input sequence packing~\citep{ding2024fewertruncationsimprovelanguage}, merging shorter sequences into a single context window of up to 4096 tokens. Both the beginning-of-sequence and end-of-sequence delimiters were set to \texttt{<|endoftext|>}, as defined by the Qwen tokenizer. Packing was performed using a linear scaling algorithm prior to training. For 
computational efficiency, we used a standard full causal attention mask (as in~\citep{deepseekai2025deepseekv3technicalreport}), rather than a multi-segment lower-triangular mask~\citep{poedator4dmasks2024}, which would otherwise require quadratic memory to store the mask itself, as well as the inability to use high-speed attention kernels.

\subsection{Training Infrastructure and Parallelization Strategy}

Training was implemented in Python 3.12 using PyTorch 2.7.1 and CUDA 12.6, with the SDPA backend for efficient attention computation on NVIDIA A100 GPUs. All experiments were conducted using the HuggingFace Transformers library version 4.52.3, Accelerate 1.7.0, and PEFT 0.15.2.

Training was performed on the Komondor HPC cluster, Hungary's national supercomputing facility at the University of Debrecen~\citep{komondor2025docs}. We utilized the GPU-accelerated partition, comprising nodes equipped with 4$\times$NVIDIA A100 GPUs (40\,GB) and 64-core AMD EPYC 7763 CPUs, interconnected via HPE Slingshot. 

Given the 4B parameter size of our model, which fits within the 40\,GB memory of a single A100 GPU using gradient checkpointing, we empirically compared Fully Sharded Data Parallel (FSDP) and Distributed Data Parallel (DDP) strategies. While FSDP reduces per-GPU memory usage by sharding model states, it incurs frequent, latency-sensitive \texttt{all-gather} operations, which proved to be a bottleneck on Komondor's interconnect. In contrast, DDP replicates the model on each GPU and synchronizes gradients via a single, bulk \texttt{all-reduce} after each backward pass, resulting in less frequent communication. 

We conducted performance measurements on a distributed setup consisting of two nodes, each equipped with four GPUs. In DDP configuration, the training throughput reached $2.9$–$3.0$ seconds per iteration. In contrast, FSDP configuration delivered substantially lower performance, reaching $5.0$–$5.1$ seconds per iteration. Mixed-precision training with the standard FSDP implementation is not possible, training with full precision resulted in OOM for our minimal training context length of $4$k tokens. As DDP yielded superior scaling and GPU utilization, we adopted it for all experiments as a hardware-aware choice for parallelization on the available infrastructure.

Training was performed on $64$ A100 GPUs ($16$ nodes) with a per-device batch size of $2$ throughout $287$ hours resulting in a total GPU time of $2.1$ years. During training we experienced $1$ unexpected (hardware failure) and $2$ scheduled stops which resulted in no significant loss of compute with checkpoints in place. The estimated carbon emission for the training is $1290\text{ kg CO}_2\text{ eq.}$ which is a below average emission due to the environmentally friendly design of the HPC utilized.

\begin{table}[h!]
\centering
\caption{Tokenizer performance on the validation set: Subword fertility (lower is better) and relative change from Qwen3-4B.}
\label{tab:tokenizer-results}
\setlength{\tabcolsep}{6pt} 
\begin{tabular}{lccc}
\toprule
\textbf{Language} & \textbf{Qwen3-4B} & \textbf{\racka{}-4B} & \textbf{Change (\%)} \\
\midrule
\textbf{Hungarian} & 3.1269 & 1.6584 & -46.96 \\
\textbf{English}   & 1.5705 & 1.9387 &  23.44 \\
\textbf{German}    & 2.0502 & 2.3090 &  12.62 \\
\bottomrule
\end{tabular}
\end{table}

\begin{table}[ht]
\centering
\caption{Results on the HULU, OpenHuEval and LM-Eval-Harness benchmarks. HULU results are measured on the publicly available validation set as an average of multiple runs taking the better option from the LoRA and full finetuned models. In case of Qwen and Racka models we patch the original OpenHuEval implementation to be compatible with these models. For details see Appendix~\ref{app:huwildbench}. LM-Eval-Harness was evaluated with and without chat templates and the best overall result is kept (with chat template for \racka{} and without it for the other models). The best result per benchmark is highlighted with \textbf{bold}.}
\label{tab:hulu}
\setlength{\tabcolsep}{7pt}
\resizebox{1.0\textwidth}{!}{%
\begin{tabular}{l|ccc|ccc|ccc|ccc}
\hline
\multicolumn{13}{c}{HULU benchmark}  \\
\hline
 \textbf{Dataset} & \multicolumn{3}{c|}{\textbf{Qwen3-4B}} & \multicolumn{3}{c|}{\textbf{\racka{}-4B}} & \multicolumn{3}{c|}{\textbf{Qwen3-4B-Base}} & \multicolumn{3}{c}{\textbf{PULI-LlumiX-Llama-3.1 8B}} \\
 & ACC & MCC & F1 & ACC & MCC & F1 & ACC & MCC & F1 & ACC & MCC & F1 \\
\hline
\textbf{HuCOLA} & \num{0.8109} & \num{0.3482} & \num{0.7840} & \num[math-rm=\mathbf]{0.8624} & \num{0.5657} & \num{0.8563} & \num{0.8254} & \num{0.4044} & \num{0.8027} & \textbf{\num{0.8989}} & \textbf{\num{0.6920}} & \textbf{\num{0.8969}}	 \\
\textbf{HuCOPA} & \num{0.5589} & \num{0.1181} & \num{0.5584} & \num{0.7990} & \num{0.5998} & \num{0.7988} & \num{0.5845} & \num{0.1705} & \num{0.5837} & \textbf{\num{0.9359}} & \textbf{\num{0.8720}} & \textbf{\num{0.9359}} \\
\textbf{HuSST} & \num{0.7517} & \num{0.5022} & \num{0.7433} & \num{0.7603} & \num{0.5137} & \num{0.7511} & \num{0.7539} & \num{0.5082} & \num{0.7513} & \textbf{\num{0.7804}} &\textbf{ \num{0.5598}} & \textbf{\num{0.7698}} \\
\textbf{HuRTE} & \textbf{\num{0.9078}} & \textbf{\num{0.8142}} & \textbf{\num{0.9078}} & \num{0.8790} & \num{0.7553} & \num{0.8790} & \num{0.8872} & \num{0.7719} & \num{0.8872} & \num{0.8979} & \num{0.7936} & \num{0.89769} \\
\textbf{HuWNLI} & \num{0.5033} & \num{-0.098} & \num{0.3862} & \textbf{\num{0.5666}} &\textbf{ \num{0.1031}} & \textbf{\num{0.4548}} & \num{0.5366} & \num{-0.060} & \num{0.4069} & \num{0.38} & \num{-0.2815} & \num{0.3668} \\
\textbf{HuCommitmentBank} & \textbf{\num{0.7378}} & \textbf{\num{0.6078}} & \textbf{\num{0.7316}} & \num{0.6388} & \num{0.4741} & \num{0.6373} & \num{0.6291} & \num{0.4733} & \num{0.6112} & \num{0.4854} & \num{0.2742} & \num{0.4594} \\
\hline
\textbf{Overall Performance (HULU)} & \num{0.711} & \num{0.382} & \num{0.685} & \textbf{\num{0.751}} & \textbf{\num{0.502}} & \textbf{\num{0.7295}} & \num{0.702} & \num{0.378} & \num{0.673} & \num{0.729} & \num{0.485} & \num{0.721} \\
\hline
\multicolumn{13}{c}{OpenHuEval benchmark}  \\
 \hline
    \textbf{HuWildBench} (WBScore) & \multicolumn{3}{c|}{\textbf{\num{63,03}}} & \multicolumn{3}{c|}{\num{57,17}} & \multicolumn{3}{c|}{\num{52,59}} & \multicolumn{3}{c}{\num{17,77}} \\ 
    \textbf{HuSimpleQA} (Acc) & \multicolumn{3}{c|}{\num{7,30}} & \multicolumn{3}{c|}{\num{10.05}} & \multicolumn{3}{c|}{\num{5,91}} & \multicolumn{3}{c}{\textbf{\num{20.03}}} \\
    \textbf{HuProverbRea} (Acc OE) & \multicolumn{3}{c|}{\num{62,47}} & \multicolumn{3}{c|}{\num{61.94}} & \multicolumn{3}{c|}{\num{41,15}} & \multicolumn{3}{c}{\textbf{\num{75.86}}} \\
    \textbf{HuProverbRea} (Acc 2CQ) & \multicolumn{3}{c|}{\num{74.98}} & \multicolumn{3}{c|}{\textbf{\num{77,53}}} & \multicolumn{3}{c|}{\num{0}} & \multicolumn{3}{c}{\num{77.36}} \\
    \textbf{HuMatchingFIB} (B Acc) & \multicolumn{3}{c|}{\num{39.59}} & \multicolumn{3}{c|}{\num{38.93}} & \multicolumn{3}{c|}{\textbf{\num{42,3}}} & \multicolumn{3}{c}{\num{33.54}} \\
    \textbf{HuMatchingFIB} (Q Acc) & \multicolumn{3}{c|}{\textbf{\num{5.94}}} & \multicolumn{3}{c|}{\num{4.68}} & \multicolumn{3}{c|}{\num{5,58}} & \multicolumn{3}{c}{\num{3.96}} \\
    \textbf{HuStandardFIB} (B Acc) & \multicolumn{3}{c|}{\num{13.20}} & \multicolumn{3}{c|}{\num{18.98}} & \multicolumn{3}{c|}{\num{0}} & \multicolumn{3}{c}{\textbf{\num{29.16}}} \\
    \textbf{HuStandardFIB} (Q Acc) & \multicolumn{3}{c|}{\num{1.08}} & \multicolumn{3}{c|}{\textbf{\num{2.15}}} & \multicolumn{3}{c|}{\num{0}} & \multicolumn{3}{c}{\textbf{\num{2.15}}} \\ \hline
    \textbf{Overall Performance (OpenHuEval)} & \multicolumn{3}{c|}{\num{33,44}} & \multicolumn{3}{c|}{\textbf{\num[round-precision = 2]{33.92875}}} & \multicolumn{3}{c|}{\num{18,44}} & \multicolumn{3}{c}{\num{32,47}} \\ \hline
\multicolumn{13}{c}{LM-Eval-Harness benchmark}  \\
    \hline
    \textbf{Arc\_hu} (Acc) & \multicolumn{3}{c|}{\num{0.3202}} & \multicolumn{3}{c|}{\num{0.3450}} & \multicolumn{3}{c|}{\num{0.3792}} & \multicolumn{3}{c}{\textbf{\num{0.3861}}} \\
    \textbf{Arc\_hu} (Acc\_norm) & \multicolumn{3}{c|}{\num{0.3844}} & \multicolumn{3}{c|}{\num{0.4101}} & \multicolumn{3}{c|}{\num{0.4169}} & \multicolumn{3}{c}{\textbf{\num{0.4323}}} \\
    \textbf{Hellaswag\_hu} (Acc) & \multicolumn{3}{c|}{\num{0.3369}} & \multicolumn{3}{c|}{\num{0.3656}} & \multicolumn{3}{c|}{\num{0.3610}} & \multicolumn{3}{c}{\textbf{\num{0.4241}}} \\
    \textbf{Hellaswag\_hu} (Acc\_norm) & \multicolumn{3}{c|}{\num{0.4095}} & \multicolumn{3}{c|}{\num{0.4510}} & \multicolumn{3}{c|}{\num{0.4557}} & \multicolumn{3}{c}{\textbf{\num{0.5606}}} \\
    \textbf{MMLU\_hu} (Acc) & \multicolumn{3}{c|}{\num{0.5427}} & \multicolumn{3}{c|}{\num{0.5378}} & \multicolumn{3}{c|}{\textbf{\num{0.5965}}} & \multicolumn{3}{c}{\num{0.5310}} \\
    \textbf{TruthfulQA\_hu\_mc1} (Acc) & \multicolumn{3}{c|}{\num{0.3177}} & \multicolumn{3}{c|}{\textbf{\num{0.3644}}} & \multicolumn{3}{c|}{\num{0.3281}} & \multicolumn{3}{c}{\num{0.3035}} \\
    \textbf{TruthfulQA\_hu\_mc2} (Acc) & \multicolumn{3}{c|}{\num{0.5102}} & \multicolumn{3}{c|}{\textbf{\num{0.5493}}} & \multicolumn{3}{c|}{\num{0.5045}} & \multicolumn{3}{c}{\num{0.4883}} \\ 
    \textbf{GSM8K\_hu} (Strict-match) & \multicolumn{3}{c|}{\num{0.6330}} & \multicolumn{3}{c|}{\num{0.5299}} & \multicolumn{3}{c|}{\textbf{\num{0.6398}}} & \multicolumn{3}{c}{\num{0.47611}} \\
    \textbf{GSM8K\_hu} (Flexible extract) & \multicolumn{3}{c|}{\num{0.6285}} & \multicolumn{3}{c|}{\num{0.5329}} & \multicolumn{3}{c|}{\textbf{\num{0.6421}}} & \multicolumn{3}{c}{\num{0.4791}} \\ \hline
    \textbf{Overall Performance (LM-Eval-Harness)} & \multicolumn{3}{c|}{\num{0.453}} & \multicolumn{3}{c|}{\num{0,454}} & \multicolumn{3}{c|}{\textbf{\num{0.4805}}} & \multicolumn{3}{c}{\num{0.4546}} \\
    \hline
\end{tabular}
}
\end{table}

\subsection{Evaluation Protocol}
\label{sec:evaluation}
We assess the model from two perspectives: intrinsic language modeling capabilities and downstream Hungarian benchmark performance.


\textbf{Intrinsic Evaluation:} We use perplexity (PPL) as our primary intrinsic metric. Perplexity measures the model's uncertainty in predicting the next token in a sequence; a lower perplexity score indicates a better fit to the data distribution~\citep{10.1121/1.2016299}. We calculated perplexity on the held-out validation set containing a mix of Hungarian, English, German and code to monitor learning progress and assess for catastrophic forgetting.

\textbf{LM-Eval-Harness benchmark:}
 To measure the model's practical utility without task-specific training, we conducted few-shot evaluations on a selection of Hungarian language understanding tasks~\citep{eval-harness}. For this we utilized the LM-Eval-Harness with Hungarian translated benchmarks. We used the already implemented benchmarks ARC\_HU, MMLU\_HU, Hellaswag\_HU and TruthfulQA\_HU from LM-Eval-Harness, which is the work of ~\citep{okapi}. Additionally, we manually added the Hungarian translated version of GSM8K~\citep{cobbe2021gsm8k}, which is a mathematical reasoning dataset and was transletd by ~\citep{thellmann2024crosslingual}, to check the model reasoning capabilities on mathemical problems.
 We ran all benchmarks separately, using the default config parameters with and without chat templates as defined in the original LM-Eval-Harness repository\footnote{\url{https://github.com/EleutherAI/lm-evaluation-harness}}. Table \ref{tab:hulu} shows our results.

\textbf{HULU Benchmark}
The HULU benchmark offers a comprehensive set of supervised tasks covering a wide range of linguistic tasks, including acceptability, causal reasoning, sentiment classification, pronoun resolution, factual commitment, and reading comprehension. For the evaluation, we separately fine-tuned our model on each HULU subtask using a custom fork of the official benchmarking script\footnote{\url{https://github.com/nytud/HuLU}}. Both full and LoRA mode finetuning was performed with the following key parameters: $lora\_rank = 64$, $lora\_alpha = 128$, $lora\_dropout = 0.1$, and $batch\_size = 2$. We report the best mode as detailed in Appendix~\ref{app:hulu_details}.

\textbf{OpenHuEval benchmark}
To further evaluate the ability of the model to perform Hungarian reading comprehension and generation tasks, we conducted fine-tuned downstream evaluations using the \textsc{OpenHuEval} benchmark suite~\citep{openhueval}. This collection focuses on cloze-style question answering and comprehension tasks designed to capture various aspects of contextual understanding, reasoning, and lexical inference in Hungarian.
OpenHuEval comprises five datasets: HuWildBench (real Hungarian user queries), HuSimpleQA (fact-based Hungarian questions), HuProverbRea (reasoning with Hungarian proverbs), HuMatchingFIB (option-based fill-in-the-blank), and HuStandardFIB (free-form fill-in-the-blank).
The evaluation was made using the original repository of the benchmark\footnote{\url{https://github.com/opendatalab/OpenHuEval}}. 

\section{Results and Analysis}
\label{sec:results}

To measure the performance of our methodology, in this section we analyze both the tokenizer and the Hungarian performance of \racka{}-4B. Benchmarks include tasks related to base language use, natural language inference, logic and Hungarian-specific knowledge. We compare our model to the original Qwen3-4B model, the vanilla Qwen3-4B-Base language model and a state-of-the-art Hungarian model, Puli-LlumiX-Llama-3.1 8B. The OTP models are not openly available, benchmarking them is not currently possible. Extending the benchmarks with international models, which according to our experiences tend to perform lower than models trained in Hungary, is matter of future works.

\subsection{Tokenizer Performance}
The primary goal of our tokenizer adaptation was to improve the encoding efficiency for Hungarian without degrading performance on the high-resource languages present in the base model. We evaluated this using \textit{subword fertility}, which is the average number of tokens required to represent a single whitespace delimited word \citep{acs-etal-2021-evaluating}.

The results measured on the validation documents, presented in Table \ref{tab:tokenizer-results}, confirm the success of our adaptation strategy. For Hungarian, the \racka{} tokenizer demonstrates a substantial improvement over the original Qwen3 tokenizer. Subword fertility was reduced by over $46\%$, from $3.13$ down to $1.66$. This means that, on average, the same Hungarian text can be represented with almost half of the original Qwen tokens, leading to direct, in most cases linear improvements in inference efficiency and latency. Crucially, this improvement in Hungarian was achieved with only modest trade-offs for other languages: fertility increased by less than 13\% for German and 24\% for English. This indicates that our vocabulary extension successfully preserved the efficiency for high-resource languages.

\subsection{Continual Pretraining Results}
Following the successful tokenizer adaptation, we successfully conducted the continual pretraining phase on the full 160B token dataset. The training ran for $326\,357$ steps on $64$ GPUs in parallel, with a per-device batch size of $2$ and gradient accumulation set to $4$, resulting in an effective logical batch size of $512$ sequences with a context length of $4096$ for each model update.
The training loss curve is depicted in Appendix~\ref{app:curve}. We see that even though the training loss begins to flatten after $50\,000$ steps, the validation perplexity continues to steadily decrease to a final value of $6.99$, indicating that overfitting is not reached.

The HuLU benchmark paints a positive picture (see the top section of Table~\ref{tab:hulu}) with a top overall performance. \racka{} seems to be behind the base model(s) on HuRTE and HuCommitmentBank, but outperforms them on the other four tasks. On HuWNLI and HuCommitmentBank, it even surpasses the current state-of-the-art 8B Puli model and achieves the best result out of all tested models on the former\footnote{All numbers are measured on the validation set.}.


The middle section of Table~\ref{tab:hulu} presents the results of evaluating \racka{} on OpenHuEval. We also denote that larger ($7$-$8$ B) models in the original OpenHuEval~\cite{openhueval} paper have an overall score of $28.77$ (LLama-3.1 8B) and $25.64$ (Qwen2.5-instruct 7B). The results show that \racka{} delivers competitive performance relative to models twice its size and to the corresponding base models with a higher overall score. \racka{} performs well in tasks that rely on interpretation and reasoning and a lower score on infilling and QA tasks.

Finally, we evaluate \racka{} on tasks from the LM-Eval-Harness Hungarian benchmark. The results are presented at the bottom of Table~\ref{tab:hulu}. On this benchmark the overall performance of \racka{} is on par with the original Qwen3-4B and the larger Puli model, with Qwen3-4B-Base leading the benchmarks.
The results indicate that \racka{} exhibits reduced mathematical reasoning capabilities compared to the base model, but demonstrates improved performance on question-answering tasks. Notably, \racka{} outperforms PULI on TruthfulQA, contrasting with its lower performance on HuSimpleQA. We attribute this discrepancy to the broader, less culture-specific nature of TruthfulQA questions, and the limitations of \racka{}'s culture-specific fact memorization capacity compared to a model twice its size.

\section{Conclusion and Future Directions}
\label{sec:conclusion}
This work introduced \racka{}-4B, a resource-efficient, continually pretrained language model for Hungarian, using a practical methodology specifically designed for deployment on the national Komondor HPC infrastructure. Through targeted tokenizer adaptation by expanding and initializing the vocabulary to better capture Hungarian specific vocabulary elements, we achieved a substantial reduction in subword fertility, improving both computational efficiency and the representational capacity of the model. The construction of a 160B-token multilingual corpus, with a carefully balanced data mixture, further enabled robust language adaptation while reducing catastrophic forgetting of high-resource languages.

The evaluation results confirm the effectiveness of our tokenizer adaptation and training strategy. With the generation speed improvements outlined above, the overall performance of \racka{}-4B is slightly higher than the base Qwen3-4B models, in some cases topping 8B models such as the latest Puli.

Future work includes systematic hyperparameter optimization, extended benchmarking and supervised fine-tuning for downstream applications as well as preference tuning. We openly publish our model and tokenizer, to support community-driven research and the advancement of digital sovereignty of Hungary.

\ifx\anonym\undefined
\section*{Acknowledgement}

We would like to thank Levente Szabados for the name idea and initial informal discussions.



This research was supported by the EKÖP-24 University Excellence Scholarship Program of the Ministry for Culture and Innovation, funded by the National Research, Development and Innovation Fund.

The authors acknowledge the support of the National Laboratory for Digital Heritage. Project no. 2022-2.1.1-NL-2022-00009 has been implemented with the support provided by the Ministry of Culture and Innovation of Hungary from the National Research, Development and Innovation Fund, financed under the 2022-2.1.1-NL funding scheme.



We acknowledge the Digital Government Development and Project Management Ltd. for awarding us access to the Komondor HPC facility based in Hungary.
\fi

\bibliographystyle{splncsnat_en}
\bibliography{sn-bibliography}

\clearpage

\section*{Appendix}
\renewcommand{\thesubsection}{\Alph{subsection}}

\subsection{Tokenizer Adaptation Algorithm}
\label{app:huntok}  

\begin{algorithm}[h]
\caption{Extended Tokenizer Construction for Hungarian Adaptation}
\label{alg:hun-tokenizer}
\begin{algorithmic}[1]
\Require Original vocabulary $\mathcal{V}$ with size $m$; Hungarian corpus $\mathcal{C}$; target increment $n=32{,}000$; target size $m^\star = m + n$
\Ensure Extended vocabulary $\mathcal{V}_{\mathrm{ext}}$ of size $m^\star$
\State Train a BPE tokenizer of size $m^\star$ on $\mathcal{C}$ to obtain the ordered vocabulary $\mathcal{V}' = (p_0, p_1, \dots, p_{m^\star-1})$.
\State Let $k$ be the index of the $n$-th token $p_k$ in $\mathcal{V}'$ such that $p_k \notin \mathcal{V}$.
\If{$k = m^\star - 1$} 
    \State \Return $\mathcal{V}'$
\Else
    \State $\mathcal{V}'' \gets$ all tokens in $\mathcal{V}$ not in $\mathcal{V}'[:k+1]$, ordered by $\mathcal{V}$'s original merge order
    \State \Return $\mathcal{V}'[:k+1]$ concatenated with $\mathcal{V}''$
\EndIf
\end{algorithmic}
\end{algorithm}

\newpage

\subsection{Composition of the Hungarian Corpus}
\label{app:hun_corpus}

\sisetup{
  round-precision = 2,
}%

The table below shows the composition of the Hungarian corpus with the corresponding number of documents and BPE tokens. For Common Crawl and News, the numbers shown are after filtering and deduplication. The former represent about 30\% of the pre-filtered data.

The Repositories dataset contains books and electronic journals (academic and otherwise) from the following sources:

\begin{description}
  \item[EPA] The \fhref{https://epa.oszk.hu/}{Elektronikus Periodika Adatbázis} (Database of Electronic Periodicals)
  \item[MEK] \fhref{https://www.mek.oszk.hu/hu/}{Magyar Elektronikus Könyvtár} (Hungarian Electronic Library)
  \item[OAI] Open Archives Initiative repositories
  \item[OJS] Open Journal System repositories
\end{description}
  
\begin{table*}[h]
  \centering
  \setlength{\tabcolsep}{6pt} 
  \begin{tabular}{@{\extracolsep\fill}llrrrr}
    \toprule%
      Dataset & Subset & Documents (thousands) & Tokens (millions) & Percentage \\
    \midrule
      \multirow{4}{*}{Common Crawl} & \texttt{.hu} & \num{52775.312} & \num{40621.051623} & \multirow{4}{*}{\num{72.708}\%} \\
      & \texttt{.ro} & \num{552.187} & \num{397.695476057} & \\
      & \texttt{.sk} & \num{446.774} & \num{224.604267996} & \\
      & \texttt{.com} & \num{2984.912} & \num{2641.12613456} & \\
    \midrule
      \multirow{4}{*}{Repositories} & EPA & \num{300.790} & \num{2672.37998772} & \multirow{4}{*}{\num{19.478972}\%} \\
      & Books & \num{32.830} & \num{1338.89434093} & \\
      & OJS & \num{27.702} & \num{96.3628461192} & \\
      & OAI & \num{349.398} & \num{7649.32137074} & \\
    \midrule
      \multicolumn{2}{@{}l@{}}{News} & \num{5938.789} & \num{2723.86983503} & \num{4.5}\% \\
    \midrule
      \multicolumn{2}{@{}l@{}}{Court} & \num{198.296} & \num{1110.3043573} & \multirow{4}{*}{\num{3.3}\%} \\
      \multicolumn{2}{@{}l@{}}{HuParl} & \num{1.707} & \num{140.560154724} & \\
      \multicolumn{2}{@{}l@{}}{OpenSubtitles} & \num{88.519} & \num{508.851908561} & \\
      \multicolumn{2}{@{}l@{}}{Wikipedia} & \num{158.457} & \num{232.159235234} & \\
    \midrule
      \multicolumn{2}{@{}l@{}}{Sum} & \num{63855.673} & \num{60357.1815379} & \num{100}\% \\
    \bottomrule
  \end{tabular}
\end{table*}

\newpage
\subsection{Fixing OpenHuEval for Qwen3-like Models}
\label{app:huwildbench}  

We have encountered several difficulties running OpenHuEval. The prompts and configurations contain minor mistakes and reasoning models, such as Qwen3, are not always evaluated correctly. Some of these problems are rooted in the OpenCompass framework itself. The most glaring issue was the omission of the OpenHuEval example scripts assumed to be present by the documentation. This was remedied by the authors after we brought it to their attention.

We have also found that the OpenHuEval tasks handled the reasoning output of models inconsistently. Some, such as HuMatchingFIB, detected the presence of the reasoning tokens and removed them before evaluation; some, such as HuSimpleQA, did not. In many cases, the presence of reasoning traces resulted in a parsing error or the model running out of the allotted context. By default, we wanted to disable reasoning in these benchmarks as small reasoning models are prone to getting stuck in thinking loops when running them with temperature $0$, which is needed for reproducibility. This behavior is also outlined in the Qwen3 documentation.

Due to time constraints, we decided to fix these issues locally, with as little change as possible in the OpenCompass codebase: we modified the chat template to respect the \texttt{enable\_thinking} tokenizer argument in the model configuration and disabled reasoning for the affected tasks. We made sure to leave the benchmark's datasets unchanged. The only exception was the HuWildBench task, which posed additional challenges that had to be addressed.

\begin{enumerate}
    \item \textbf{Remove artifacts present in the Hungarian prompts only.} The HuWildBench prompt is a user prompt that is segmented into three logical units. An instruction, a question and a description. These are not delimited explicitly in the English or Chinese prompts (also provided as part of the HuWildBench dataset), however the Hungarian prompt contains the following phrases as separators attached to the end of the previous field directly without even a whitespace: ``a kérdés az: '', ``a leírás: '', ``a válasz: '' for the question, description and the indication of the answer starting point respectively. We believe that these unclear separators put directly as part of the last word erode the performance of language models, especially smaller ones. Thus we modify the prompts to keep the separators but we add two empty lines between the last character of the previous unit and the separator. We also improve the capitalization and fluency of these phrases while removing the answer start indicator as we will use the chat prompt template to denote assistant turns. This results in the following separators: ``\textbackslash{}n\textbackslash{}nA kérdés: '', ``\textbackslash{}n\textbackslash{}nA leírás: ''.
    \item \textbf{Use the proper prompt template.} The Qwen3 and Racka models use a chat prompt template with system, user and assistant message frames and are trained to follow instructions in this format. Thus, we attempted to apply the default chat template of these models in OpenCompass which setting was not respected. We opted to modify the implementation to include chat templates in HuWildBench. As Qwen3 and derivative models also need a system prompt we add a modification of the industry default ``You are a helpful assistant.'' adding the language constraint as ``You are a helpful Hungarian assistant.''. For the non-instruct Qwen3-4B-Base and PULI-LlumiX-Llama-3.1 (the latter we tested with the standard Llama-3.1 chat template which yielded slightly wores results) models we add the answer trigger prompt as ``\textbackslash{}n\textbackslash{}nA válasz: '' and do not use any special prompt template.
    \item \textbf{Increasing model generation control.} As mentioned above, we make sure that the tokenizer settings are respected to avoid generating reasoning traces. We also enforce repetition penalty and frequency penalty using the definitions of vLLM~\cite{Kwon:2023}, which are both set to $0.3$ in the case of all Qwen and Racka models to avoid repetitive text generation. 
    \item \textbf{Fixing judge model errors.} HuWildBench is one of the tasks where the original implementation fails to remove (often English) reasoning traces from the judge model prompt. We reimplement this behavior correctly to align with the original intent present in the benchmark configuration. We also find that the judge model rarely follows the JSON schema defined in its prompts. In some cases this results in the OpenCompass implementation missing the judge score. We use a simple regex pattern to find the last ``score:'' in the ill-formed judge outputs and extract the first integer that follows it as a score. This method is able to correctly handle $99.91\%$ of the failed judge prompts. When no ``score:'' segment is present in the judge output use a default score of $1$. We enforce the usage of the single model scoring prompt to make sure OpenCompass does not automatically switch to the pairwise evaluation where a parent/base model is present.
\end{enumerate}

While HuWildBench was the only subset where these errors fully prevented the original implementation from working, we believe that future work should revisit the prompts and implementation of OpenHuEval to ensure language quality and compatibility with chat and reasoning language models.

\newpage

\subsection{HULU Downstream Training Details}
\label{app:hulu_details}

We average results over multiple trainings to control for random factors during downstream training. The model is trained with early stopping for $10$ fine tuning rounds on HuCOPA, HuWNLI and HuCommitmentBank and $5$ trainings on the other subsets.

We also perform both LoRA and full fine-tuning and report the superior method in Table~\ref{tab:hulu}. Here we detail the best method for each model and benchmark subset in Table~\ref{tab:appendix_hulu_extras}.

\begin{table}[]
    \caption{Details on which fine-tuning method is superior for each model-task combination.}
    \centering
    \begin{tabular}{l|c|c|c|c}
    \hline
        HULU subset & Qwen3-4B &\racka{}-4B & Qwen3-4B-Base & Puli-LlumiX-Llama-3.1 8B\\
        \hline
         \textbf{HuCOLA} & LoRA & LoRA & LoRA & LoRA\\
\textbf{HuCOPA}  & LoRA & Full & LoRA & LoRA\\
\textbf{HuSST}  & LoRA & LoRA & LoRA & LoRA\\
\textbf{HuRTE}  & LoRA & LoRA & LoRA & LoRA\\
\textbf{HuWNLI}  & Full & Full & Full & LoRA\\
\textbf{HuCommitmentBank} & Full & Full & Full & LoRA\\
\hline
    \end{tabular}
    \label{tab:appendix_hulu_extras}
\end{table}

\newpage
\subsection{Loss Curves}
\label{app:curve}

\begin{figure}
    \centering
    \includegraphics[width=0.85\linewidth]{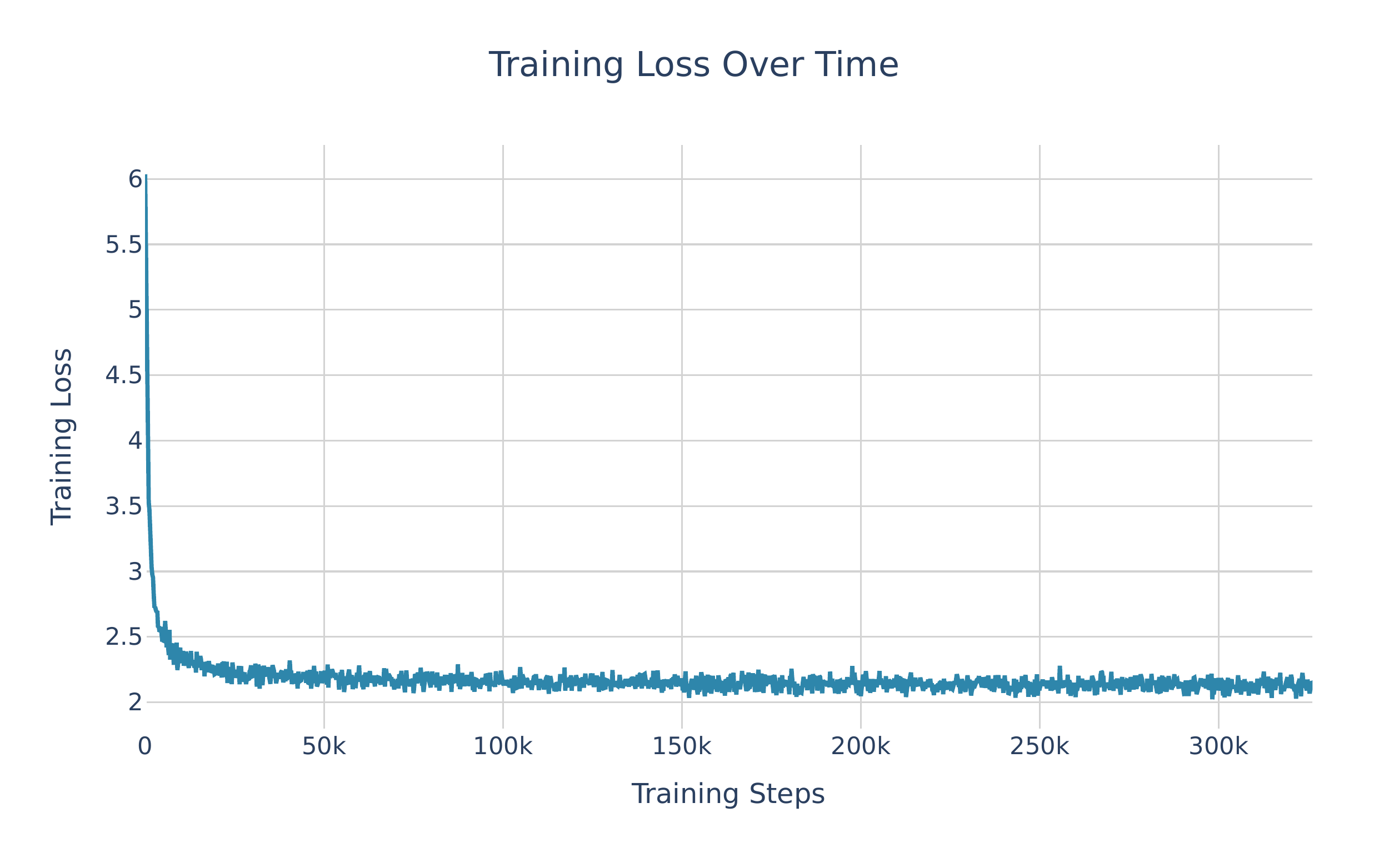}
    \includegraphics[width=0.85\linewidth]{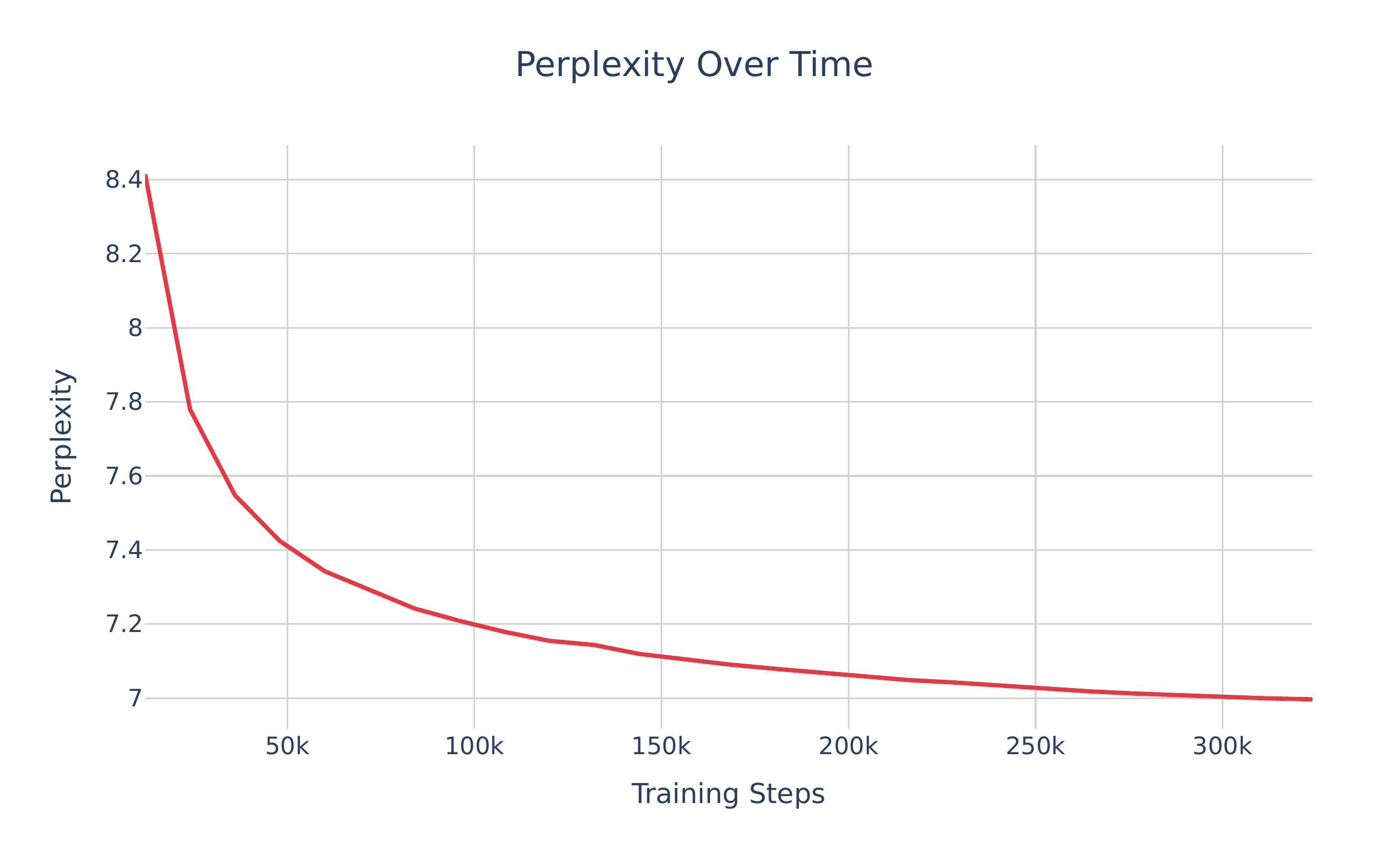}
    \caption{Average training Categorical-Crossentropy loss logged every $200$ steps (top) and validation perplexity measured every $12\ 000$ steps on a language-stratified held out set of documents (bottom).}
    \label{fig:losses}
\end{figure}

\end{document}